\documentclass[twoside,11pt]{article}
\usepackage{jair, theapa, rawfonts}
\usepackage[T1]{fontenc}
\usepackage[utf8]{inputenc}
\usepackage{microtype}
\usepackage{latexsym}
\usepackage{graphicx}
\usepackage{amsmath} 
\usepackage{amsfonts}
\usepackage{amssymb}
\usepackage{amsthm}
\usepackage{mathtools}
\usepackage{algorithm, algorithmic}
\usepackage{booktabs}
\usepackage{microtype}

\jairheading{78}{2023}{269-285}{10/2022}{10/2023}
\ShortHeadings{Dense Retrieval Knowledge Graph Augmentation}
{Abaho, \& Alfaifi}
\firstpageno{269}

\begin{document}

\title{Select and Augment: Enhanced Dense
Retrieval Knowledge Graph Augmentation}

\author{\name Micheal Abaho \email micheal.abaho@liverpool.ac.uk \\
      \addr University of Liverpool, United Kingdom \\
       \AND
       \name Yousef H. Alfaifi \email y\_alfaifi@ut.edu.sa \\
       \addr Faculty of Computers and Information Technology,\\University of Tabuk, Tabuk, Saudi Arabia}

\maketitle

\begin{abstract}
Injecting textual information into knowledge graph (KG) entity representations has been a worthwhile expedition in terms of improving performance in KG oriented tasks within the NLP community. External knowledge often adopted to enhance KG embeddings ranges from semantically rich lexical dependency parsed features to a set of relevant key words to entire text descriptions supplied from an external corpus such as wikipedia and many more. Despite the gains this innovation (Text-enhanced KG embeddings) has made, the proposal in this work suggests that it can be improved even further. Instead of using
a single text description (which would not sufficiently represent an entity because of the inherent lexical ambiguity of text), we propose a multi-task framework that jointly selects a set of text descriptions relevant to KG entities as well as align or augment KG embeddings with text descriptions. Different from prior work that plugs formal entity descriptions declared in knowledge bases, this framework leverages a retriever model to selectively identify richer or highly relevant text descriptions to use in augmenting entities. Furthermore, the framework treats the number of descriptions to use in augmentation process as a parameter, which allows the flexibility of enumerating across several numbers before identifying an appropriate number. Experiment results for Link Prediction demonstrate a 5.5\% and 3.5\% percentage increase in the Mean Reciprocal Rank (MRR) and Hits@10 scores respectively, in comparison to text-enhanced knowledge graph augmentation methods using traditional CNNs.
\end{abstract}

\section{Introduction}
\label{Introduction}
Jointly learning relational information by using both textual mentions and knowledge graph (KG) mentions of entity pairs has improved performance in not just knowledge base (KB) completion tasks, such as link and relation prediction \shortcite{bordes2013translating,gardner2014incorporating}, but also in various other NLP tasks such as fact retrieval \cite{bordes2013translating} and analogical reasoning \shortcite{gentner2017analogical,mikolov2013distributed}. More so, these Text-enhanced knowledge graph embedding (KGE) methods have been recently used to enrich representations of entities in order to improve performance in domain-specific tasks such as Biomedical Named Entity Recognition \shortcite{yuan2021improving}, Medical Natural Language Inference MedNLI \shortcite{michalopoulos2020umlsbert}, and Normalising medical concepts \cite{limsopatham2016normalising}.  

\begin{figure}[t]
    \centering
    \includegraphics[width=0.65\columnwidth]{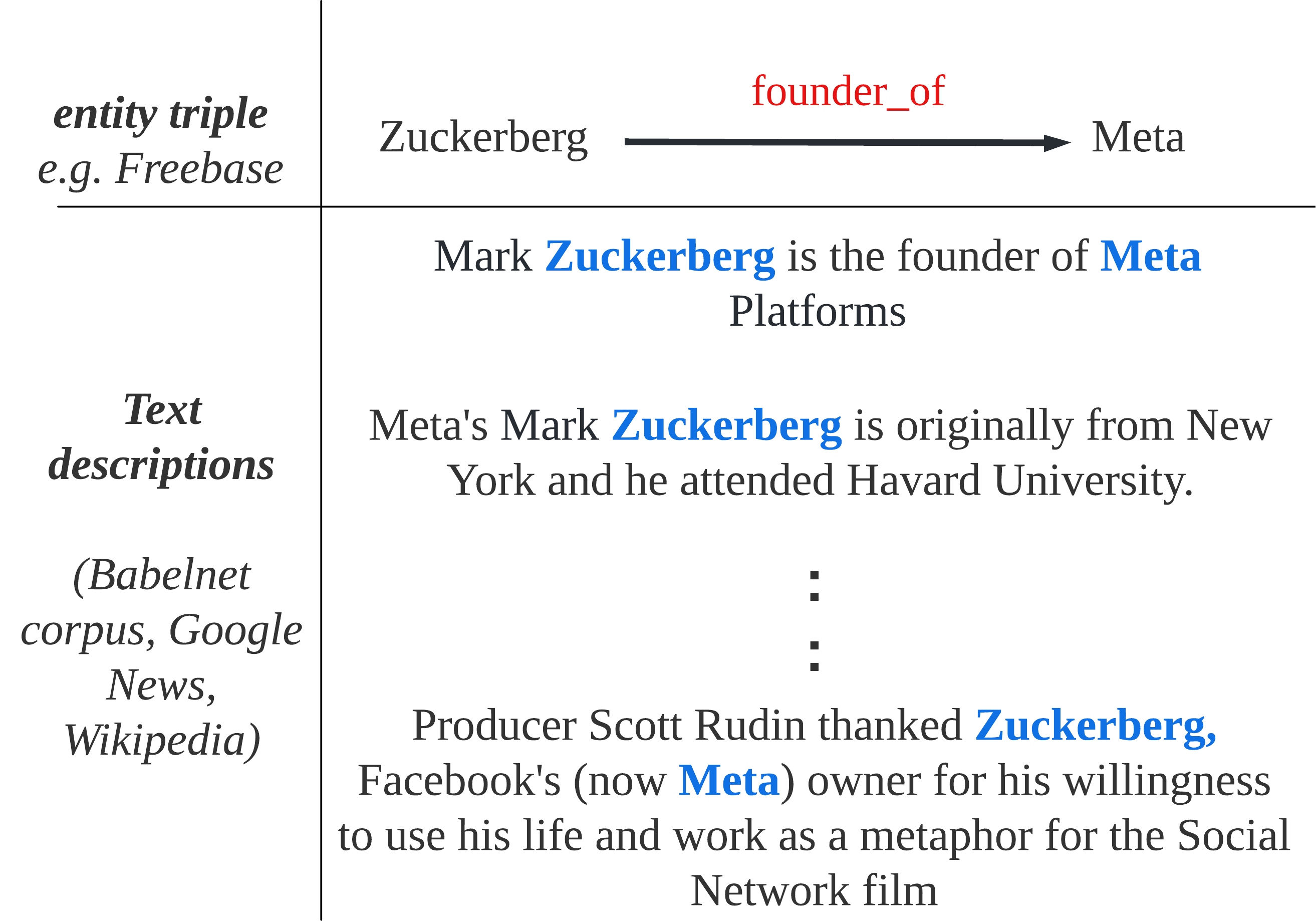}
   \caption{An example of KB triple and corresponding set of relevant text descriptions.}
    \label{fig:text_desc}
 
\end{figure}

Typically, the objective in adapting text to the KG is to maximise the similarity between vectors encoding KG entities, and vectors encoding text descriptions in which the entities are mentioned. For example, given a fact such as ``Zuckerberg is the founder of MetaPlatforms'', \shortciteA{zhong2015aligning} aligns a KGE model vector for the entity Zuckerberg to a text embedding model vector of the same entity, however obtained from a text depicting this fact. These aligned representations generated for both object and subject entities would then be encoded by a model that uses a scoring function to indicate existence or non-existence of a relationship triple between the subject and object i.e. \textit{(subject, relation, object)}. 

We notice that prior efforts aligning textual descriptions to KG’s often rely on a single text description of an entity which is often provided in the KB being studied or an external KB, such as Wikipedia \shortcite{toutanova2015representing}. More so, some works constrain a relevant text description to a sequence of words within a text window of a predefined size \shortcite{xie2016representation,zhong2015aligning}. The downside of these works is the assumption that a single description accompanying an entity in a KB would sufficiently and comprehensively represent the entity. This assumption is however, an over-estimated and exaggerated expectation of a single entity description. Even if it were a quality description of the entity, it would not contain all co-occurrences of the entity and its related entities. Moreover, word sense disambiguation research indicates that a word can have multiple glosses (word senses) depending on the context in which its used \shortcite{huang2019glossbert,blevins2020moving,scarlini2020more}.

To address this lexical ambiguity, as well as concerns around the exaggerated assumption mentioned in previous paragraph, \shortciteA{kartsaklis2018mapping} arbitrary gathers textual descriptions for an entity from numerous sources such as WordNet, Wikipedia, and FrameNet. They then use a term-based weighting mechanism (Tf-idf) for computing probabilities of an entity. Recently, \shortciteA{veira2019unsupervised} augmented KG entity embeddings using word2vec embeddings obtained by training word2vec \shortcite{mikolov2013distributed} on an external corpus that contained mentions of the KG entities. These efforts are worthy of recognition in terms of minimising nuances brought by lexical ambiguity, however, we recognize two challenges, 1) the arbitrary gathered formal descriptions of entities may not contain an entity's related entities i.e. no co-occurrence of related entities, more so, the gathering process may un-wantedly become tedious for enormous KGs, 2) they utilize traditional vectorization techniques, which also have their limitations. Tf-idf ignores properties such as word-order and co-occurrence statistics, which are necessary when generating a semantic vector representation, and word2vec learns the same vector for a word irrespective of the context in which it is used.

In this work, we propose Dense Retrieval KG Augmentation (DRKA), a multi-task framework which uses dense retrieval to obtain documents or text descriptions semantically relevant to an entity pair, and subsequently augments a given KG with dense representations. This method addresses the first challenge in preceding paragraph by introducing a retriever model which automatically selects a richer set of descriptions to use for augmenting KG embeddings. To address the second challenge, DRKA takes full advantage of transformer-based SBERT \shortcite{reimers2019sentence} to encode text descriptions. SBERT (Sentence-BERT) and other contextual language models (CLMs) have proven their superiority over traditional embedding methods in various downstream NLP tasks \shortcite{brown2020language}. Generally, DRKA builds on the idea of using multiple entity descriptions \shortcite{kartsaklis2018mapping,veira2019unsupervised} to establish a relation between a pair of entities, such as in Figure \ref{fig:text_desc}, where the multiple sentences are used to establish the relation triple \textit{(Zuckerberg, founder\_of, Meta)}. 

Similar to Document level relation extraction tasks \shortcite{zhou2021document}, DRKA leverages CLMs ability to capture interactions among distantly or remotely connected entities. Relying on multiple text descriptions for an entity, DRKA is able to increase the probability of co-occurrence of KG entity pairs within text, which inadvertently minimises the challenge of without mention entity pairs if a small text window is considered \shortcite{kartsaklis2018mapping,veira2019unsupervised}. DRKA learns to jointly embed KG mentions and textual mentions of an entity in the same embedding space. Our proposed method is evaluated on KG completion tasks (described under Section \ref{sec:eval_results}) such as link and relation prediction using Freebase FB15k dataset \shortciteA{veira2019unsupervised}. As later shown, there is approximately 6\% and 3\% percentage increase in the overall Mean Reciprocal Rank (MRR) and Hits@10 scores respectively, for the link prediction (LP) task obtained by DRKA. This is in contrast to DKRL \cite{xie2016representation}, a model that uses Continuous bag of words (CBOW) and a CNN to generate description based representations of entities. The evaluation results are indicative of the effectiveness of augmenting KG embeddings with dense contextualised representations encoded from multiple text descriptions rather than a single text description. 

\section{Related Work}
There are several works on augmenting KGs for purposes on improving performance in link prediction tasks. This section categorises related work into three areas, these include, Knowledge bases, Text-enhanced knowledge graph completion and Dense representation learning.

\label{sec:text-kg}
\paragraph{Knowledge Bases:}KBs are often adopted in distance supervised learning \shortcite{mintz2009distant} because they are incomplete, implying that they do not possess all existing knowledge for the domains they represent, or at the bare minimum, that they do not explicitly state knowledge in its basic granular form \shortcite{reschke2014event}. There have been several efforts to alleviate the bottleneck of incompleteness such as entity linking across domains-based graphs \shortcite{schneider2022decade} and integrating entities mentioned in unstructured text \shortcite{toutanova2015representing,kartsaklis2018mapping,hakami2022learning}. Another rapidly growing paradigm that has attracted a lot of attention to KBs, is Language Models as Knowledge bases (LM-as-KBs). LM-as-KBs suggests that neural language models (LMs) can be treated as suitable alternatives or at least a proxy for KBs \shortcite{petroni-etal-2019-language}. To achieve this, researchers design experiments in which they train LMs to learn to correctly answer prompts \shortcite{petroni-etal-2019-language,heinzerling2020language,gao2020making}. Quite clearly, the subject of KBs has attracted a lot of research across domains with a keen interest in knowledge representation, management and dissemination. The focus in this work is augmenting KGE representations with external text. Moreover, we aim to enhance the augmentation process by enabling selection of richer text descriptions as opposed to simply adopting formal entity descriptions that may not sufficiently represent the entity in the context of its relationship with other entities. 

\paragraph{Text-enhanced Knowledge Graph Completion:} To improve performance in tasks such as LP, several authors have undertaken efforts to augment KG embeddings using external text. \shortciteA{wang2016text} recently used co-occurrences between entities and words in text to enrich entity and relation representations in order to better handle 1-to-N, N-to-1, and N-to-N relations. They used point-wise and pairwise contexts using co-occurrence frequencies to build textual context embeddings, which were then used to enrich embeddings of the KG components. \shortciteA{kartsaklis2018mapping} extends a KG by adding Tf-idf weighted terms from textual descriptions of entities, and later uses a multi-sense LSTM to learn multi-sense embeddings in order to achieve sensitivity towards lexically ambiguous words, i.e. words having more than one disjoint meaning (homonymy) and words with multiple different meanings (polysemy). \cite{riedel2013relation} combines text-driven KG-based relations in the same entity-pair co-occurrence matrix, which are subsequently decomposed to obtain entity embeddings. \shortciteA{toutanova2015representing,hakami2022learning} use Lexicalised Dependency Paths (LDPs) obtained from sentences that co-occur in a text corpus as textual relations in a KG. 

Similar to the above works, our work incorporates external text into a KG, however, it differs form them in such a way that, instead of using a single description, it relies on multiple text descriptions when augmenting a KG entity embedding. Our work further distinctively differs from other works that have used embeddings initialised by training word2vec on a corpus \shortcite{veira2019unsupervised} in 2 ways, the first being, introduction of a description retrieval task in the augmentation process, thereby having a multi-task framework that learns to jointly select relevant descriptions as well as align these descriptions to KG embeddings. The second difference being the use of transformer-based SBERT, which provides high quality sentence embeddings that capture both the semantic and syntactic information of a sentence \shortcite{reimers2019sentence}.

\paragraph{Dense Representation Learning:}
Dense embedding models such as BERT \shortcite{devlin2018bert} have demonstrated an ability to effectively capture the semantics of words and sentences by encoding information about their neighbouring words and sentences, and the overall contexts in which they are mentioned. Recently, some works have shown that dense embeddings have surpassed term-based weighting schemes in tasks that require context retrieval from a piece of text in order to solve downstream tasks, such as Question Answering (QA) and Analogical Reasoning \shortcite{karpukhin2020dense,izacard2020leveraging}. Similar to how prior work has used these models for retrieval augmented generation to generate answers to questions in QA tasks \shortcite{izacard2020leveraging}, this work explores enhancing KG augmentation using dense retrieval of entity descriptions from a corpus of text. We integrate this retrieval task as an auxiliary task to provide dense representations corresponding to multiple text descriptions that are semantically related to an entity.
 
\section{Method}\label{sec3}
\begin{figure}[t]
    \centering
    \includegraphics[width=0.9\linewidth]{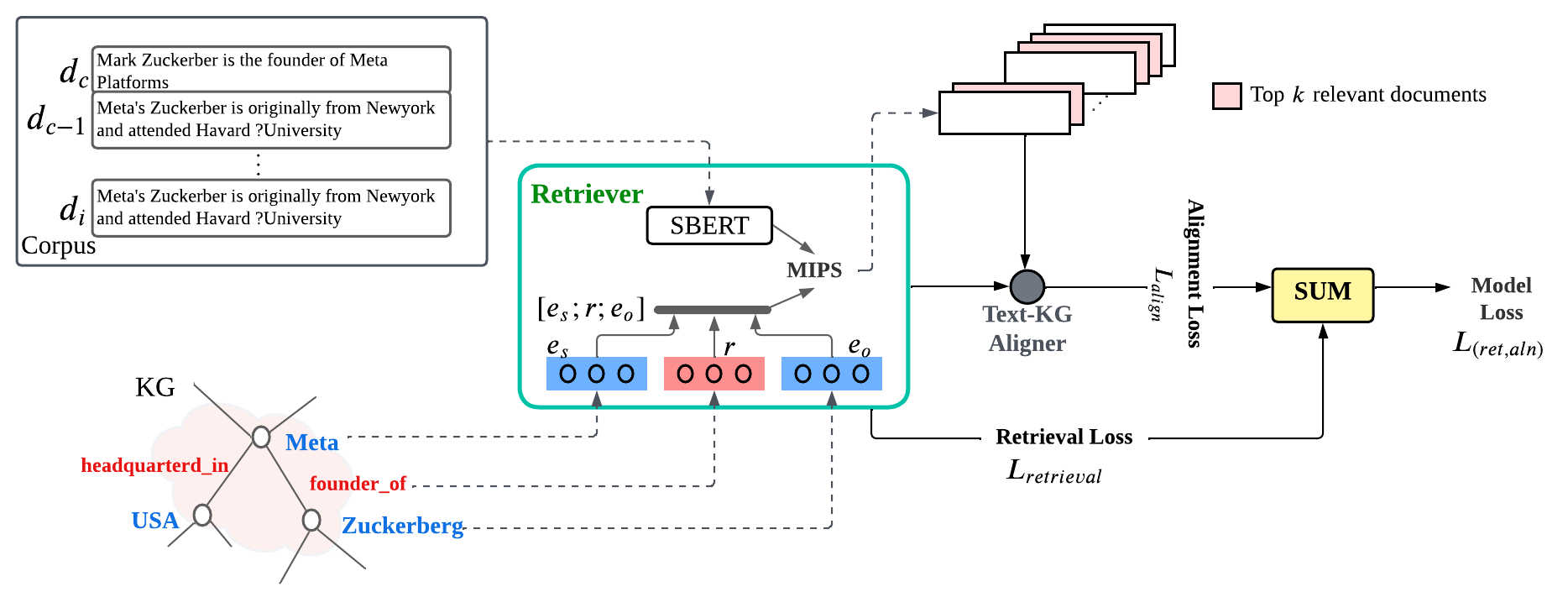}
   \caption{An overview of the proposed DRKA framework. DRKA combines a Retriever, which retrieves a set of documents (text descriptions) relevant to a relation triple and an Text-KG Aligner which fuses the identified relevant documents to KG embeddings. The last component of the framework shows a joint loss that involves summing up retrieval and alignment losses.}
    \label{fig:dkra}
        \vspace{-2em}
\end{figure}

In this work, we assume access to a KG and a corpus of documents (text descriptions). Both of these are taken as input to the DRKA framework illustrated in Figure \ref{fig:dkra}. A pre-trained SBERT model \shortcite{reimers2019sentence} is used to encode text descriptions and a KG embedding model used to encode KG entities. Maximum Inner Product Search (MIPS) aided by an attention mechanism is used for finding the top $k$ documents relevant to the query which is a concatenation of the triple elements. Concatenation is performed in order to allow knowledge transfer across elements within a triple, hence when searching for entity relevant documents, DKRA relies on knowledge of not just the entity but also its related entities.   Subsequently the identified relevant documents embeddings are aligned to the KG embeddings and the model is trained by optimising the joint loss which is a summation of the losses with respect to retrieval and alignment.

\subsection{Dense Retrieval KG Augmentation (DRKA)}
Given a graph (for a certain KB) $KG$ consisting of facts expressed in form of triples $(e_s, r, e_o)$, where $e_s$ and $e_o$ are subject (head) and object (tail) entities respectively, and $r$ is the relation linking the two together, we propose DRKA, a method that learns to cast entity embeddings and their text description embeddings into a common vector space, in order to accurately infer missing links or triples. 

Different from prior work that arbitrary choose and select descriptions to use for the entities, this work considers learning a dense retriever to automatically select top $k$ most relevant text descriptions for a particular entity triple. The idea is to maximise a LM's ability to capture missing facts from multiple sources as earlier indicated in the introduction. Similar to Dense Passage Retrieval (DPR) \shortcite{karpukhin2020dense}, this retriever computes an inner product to indicate the similarity between a dense representation of a text description and the dense representation of an entity triple. 
To incorporate textual information, we leverage pre-trained SBERT \shortcite{reimers2019sentence} to encode input text descriptions and learn to assign a similarity score between each sentence and a given query (triple), as described in Section \ref{sec:dense_dec_ret}. These scores are treated as attention scores which are used to compute an overall representation of the triple which is then aligned to the relevant text descriptions as described in Section \ref{sec:text-entity_fusion}.

\subsubsection{Dense Description Retrieval}
\label{sec:dense_dec_ret}
We initialise entities $e \in \mathcal{E}$ using TransE \shortcite{bordes2013translating} which follows the formulation $\lVert \vec{e_s} + \vec{r} - \vec{e_o} \rVert$, indicating that an object embedding $\vec{e_o}$ should be in close proximity to the summation of its corresponding subject and relation embeddings i.e. $\vec{e_s} + \vec{r}$. For a triple $e_t$, an embedding $\vec{e_t}$ is obtained through a concatenation of $[\vec{e_s};\vec{r};\vec{e_o}]$, which is then used as a query to search a corpus of external documents $C$ for documents relevant or semantically similar to $e_t$. 
To answer this query, we compute an inner product between a dense vector representation $\vec{e_t}$ of the query $e_t$, and another projected dense vector representation $\vec{d_i^{'}}$ of a description $d_i$. $\vec{d_i^{'}}$ is computed as $\vec{d_i^{'}} = W_c \vec{d_i} + b_c$, where $\vec{d_i} = \mathrm{SBERT}(d_i)$, $W_c$ is a matrix used to project the description's embeddings to the same space as the triple embedding, and $b_c$ is a bias term. Equation \eqref{eq:similarity} shows the computation for a similarity score between triple $e_t$ and description $d_i$. 

\begin{align}
\label{eq:similarity}
    sim(e_t, d_i) = \vec{e_t} \cdot \vec{d_i^{'}}^{\top}
\end{align} 

where $\vec{e_t}$ is a matrix of three vectors corresponding to the triple elements i.e. $\vec{e_t} \in \mathbb{R}^{\mid e_t \mid \times l}$, ${\mid}e_t{\mid} = 3$, $\vec{d_i^{'}} \in \mathbb{R}^{1 \times l}$, $l$ is the embedding dimension size, and $sim(e_t, d_i)$ is a vector of similarity scores between elements of the triple $e_t$ elements and description $d_i$.

\subsubsection{Text-KG Aligner}
\label{sec:text-entity_fusion}
For a triple $e_t$, we search for the $k$ most relevant documents by computing an attention score across all documents in $C$. This attention is a normalised similarity score between the documents and the triple embedding $\vec{e_t}$ as shown in  \eqref{eq:norm_similarity}). Since $sim(e_t,d_t)$ returns a vector of similarity scores, we compute its L2 norm prior to the normalisation. 

\begin{align}
\label{eq:norm_similarity}
\vec{A_t} = \frac{\exp(\lVert sim(\vec{e_t} \cdot  \vec{d_i^{'}}) \rVert)}{\sum_{\vec{d_k} \in C}\exp(\lVert sim(\vec{e_t} \cdot  \vec{d_i^{'}}) \rVert)}
\end{align}

Prior authors have benefited from generating overall representations of sentences or tokens by using attention scores to compute a weighted sum \shortcite{abaho2021detect,abaho2022position}. Inspired by these works, we generate an overall triple representation $\vec{e_t^k}$ by computing an inner product between a matrix $\vec{D_t^k}$ of top $k$ projected description embeddings relevant to triple $e_t$ i.e. $\vec{D_t^k} \in \mathbb{R}^{k \times l}$ and the attention vector $A_t$ as shown in \eqref{eq:weighted_sum},

\begin{align}
\label{eq:weighted_sum}
\vec{e_t^k} = \vec{A_t}^{\top} \cdot \vec{D_t}
\end{align} 
where $\vec{A_t} \in \mathbb{R}^{{k \times 1}}$ and $\vec{e_t^k} \in \mathbb{R}^{1 \times l}$. The attention vector scores indicate the strength of association or how relevant each document is to the triple in question.

\subsubsection{Joint retrieval and Text-KG alignment}
The proposed DRKA model is trained to jointly optimize the retrieval of relevant documents (text descriptions) as well as the alignment of the documents to a KG. During training, the trainable matrix $W_c$ used to cast text description embeddings into the same vector space as the KG entity embeddings is updated alongside the embeddings. The model is trained to minimise the loss in \eqref{eq:loss} and illustrated in Figure \ref{fig:dkra}.

\begin{align}
\label{eq:loss}
L({ret},{aln}) = L_{align} + \alpha L_{retrieval}
\end{align} 

$\alpha$ is a tunable parameter to adapt the auxiliary retrieval loss to the Text-KG alignment. We use a margin-based loss function to formulate our $L_{align}$ loss function presented in \eqref{eq:align}.

\begin{align}
\label{eq:align}
L_{align} = - \sum_{\vec{e_t^k} \in E_t}\sum_{\vec{e_t^{k^{'}}} \in E_t^{'}}max(\gamma + d(\vec{e_t^k}-\vec{e_t^{k^{'}}}),0) 
\end{align} 

where $\gamma \ge 0$, $d(\vec{e_t^k}-\vec{e_t^{k^{'}}})$ is a dissimilarity function in which we use an L1 norm, having performed well in prior knowledge graph representation approaches \shortcite{xie2016representation}. $E_t^{'}$ is a set of negative training instances drawn as shown in \eqref{eq:neg_instances}.

\begin{align}
\label{eq:neg_instances}
\begin{gathered}[b]
    E_t^{'} = \{(e_s^{'},r,e_o)\mid e_s^{'}\in\mathcal{E}\} \cup \{(e_s,r,e_o^{'})\mid e_o^{'} \in \mathcal{E}\} \\
    \cup \{(e_s,r^{'},e_o)\mid r^{'} \in \mathcal{R} \cup \{(e_s^{'},r^{'},e_o^{'})\mid r^{'} \in \{\mathcal{E},\mathcal{R}\}\} 
\end{gathered}
\end{align} 


The retrieval loss minimised is given by \eqref{eq:retrieval},
\begin{align}
\label{eq:retrieval}
L_{retrieval} = -\sum_{e_t \in E_t} \log \frac{\exp(\lVert sim(\vec{e_t} \cdot 
 \vec{d_i}) \rVert)}{\sum_{\vec{d_{i=1}}}^{{\mid}C{\mid}}\exp(\lVert sim(\vec{e_t} \cdot 
 \vec{d_i}) \rVert)}
\end{align} 

\section{Experiments}\label{sec4}
We evaluate the proposed DRKA method on three tasks: Link Prediction \shortcite{bordes2013translating}, Relation Prediction \shortcite{weston2013connecting,yao2019kg} and Triplet classification \shortcite{zhong2015aligning,yao2019kg}. We adopt the Freebase (FB15K) Knowledge graph, the Babelnet corpus \shortcite{navigli2012babelnet}, Google News dataset, and Wikipedia articles \shortcite{veira2019unsupervised}, from which we assemble text descriptions for all the KG entities.

Unlike prior authors who often eliminate entity descriptions of short lengths (or with no description at all), we do not eliminate any description during our preliminary text-preprocessing phase. We hypothesise that discarding some descriptions on account of their short length might unwittingly eliminate relevant descriptions, instead we rely on $k$ (the hyper-parameter), as discussed in Section \ref{sec:text-entity_fusion}, which if tuned well enough will enable us to obtain a sufficient number of text descriptions relevant to a triple. After pre-processing the gathered entity descriptions, the dataset is split into training, validation and test sets) and the resultant dataset statistics are presented in Table \ref{tab:table_statistics}. 

Table \ref{tab:table_statistics} provides a breakdown of the Train, Validation (Val) and Test splits used in our experiments and the ~3.8M entity descriptions which contain 96\% of the entities within the KB. As shown in the table, the text descriptions respectively cover 51.8\%, 30.4\% and 29.9\% of the Training, Validation and Testing triples. Additionally, we have an average of 5 descriptions per entity and an average of 3 descriptions per entity pair.

\paragraph{Metrics:}
\label{sec:metrics}
Following prior work on KG completion, we report two different metrics: the Mean Reciprocal Rank(MRR), and Hits@10. Percentages for MRR and Hits@10 are reported for the test sets across all experiments conducted.

\begin{table}[!t]
\begin{center}
\label{tab:table_statistics}%
\begin{tabular}{@{}lllll@{}}
\toprule
Dataset & Relations & Entities & \multicolumn{2}{l}{Train / Val / Test Triples} \\ \midrule
FB15K   & 1,341     & 14,904   & \multicolumn{2}{l}{472,860 / 57,803 / 48,991} \\
Text    & 3814190   & 14,308   & \multicolumn{2}{l}{244946 / 17572 / 14599}   \\
\bottomrule
\end{tabular}
\caption{Dataset statistics for both KB and text corpus}
\end{center}
\vspace{-1em}
\end{table}

\begin{table}[b]
\begin{center}
\begin{tabular}{@{}lcc@{}}
\toprule
\textbf{Parameter}                                                             & \textbf{Tuned-range}                                                                & \textbf{Optimal}                                   \\ \midrule
KG Embedding dimension                                                                    & {[}50,100,200,300{]}                                                                      & 200                                                 \\

$\gamma$                                                                             & {[}0.5,1.0,1.5,2.0{]}                                                                 &         1.0                                           \\
Optimizer                                                                     & {[}SGD,Adam{]}                                                                     & Adam                                               \\
Epochs                                                                             & {[}20, 50, 70, 100, 120{]}                                                                 &         70                                          \\

Learning rate                                                                  & \begin{tabular}[c]{@{}c@{}}{[}5e-4, 1e-4, 5e-3, 1e-3, \\ 5e-2, 1e-2{]}\end{tabular} & 1e-3                                               \\ \bottomrule
\end{tabular}
\label{tab:parameter}
\caption{Parameter settings for DRKA}
\end{center}
 \vspace{-2em}
\end{table}

\subsection{Baselines}
\label{sec:baseline}
Besides traditional KGE models i.e. TransE \shortcite{bordes2013translating}, DistMult \shortcite{yang2014embedding}, CompIEx \shortcite{trouillon2016complex} and RotatE \shortcite{sun2019rotate}, we use DKRL \shortcite{xie2016representation}, a text-enhanced KGE that generates entity representations by adding structured based representations, obtained from TransE, to description-based representations obtained using either CBOW or CNN Encoder. The CNN Encoder takes word2vec word embeddings as input. Additionally, we consider DRKA(DPR) which decouples the retriever from the alignment/fusion (illustrated as Text-KG aligner) in Figure \ref{fig:dkra}, in which case we formulate the triple of elements as a sentence of $(e_s,r,e_o)$ concatenated e.g. ``\textit{Zuckerberg founder of Meta}'' (i.e. to serve as a query), and train DPR to select descriptions semantically relevant to this sentence and separately train DRKA with just $L_{align}$ in Equation \ref{eq:align}. For this work, we compare variants withf the KGE's for both DKRL's architecture and DRKA’s architecture.

\begin{table}[!b]
\begin{center}
\begin{minipage}{\textwidth}
\resizebox{\linewidth}{!}{
\begin{tabular}{@{}clcccccc@{}}
\toprule
                &    & \multicolumn{2}{c}{Overall}   & \multicolumn{2}{c}{With mentions} & \multicolumn{2}{c}{Without mentions} \\ 
               &   & MRR $\uparrow$         & Hits@10 $\uparrow$       & MRR $\uparrow$             & Hits@10 $\uparrow$        & MRR $\uparrow$               & Hits@10 $\uparrow$         \\ \midrule
KG Only  & TransE              & 36.8        & 52.4 & 34.5            & \underline{54.2}            & \underline{38.2}            & \underline{58.1}            \\
& DistMult            & 36.3          & 51.8          & 34.1            & 53.3           & 36.6              & 55.7             \\
& CompIE            & \underline{37.1}          & \underline{52.8}          & \underline{34.6}            & 52.7           & 37.4              & 55.9             \\
& RotatE            & \textbf{38.8}          & \textbf{53.1}          & \textbf{35.9}            & \textbf{54.6}           & \textbf{38.6}              & \textbf{59.7}             \\
 \midrule
KG + Text & DKRL(CNN) + TransE & 38.9          & 54.1          & 38.7            & \textbf{54.5}   & \underline{40.8}              & 58.6             \\
 & DKRL(CNN) + DistMult & 37.8          & 52.9          & 37.7            & 52.4   & 39.8              & 58.4             \\
  & DKRL(CNN) + CompIE & 38.1          & 54.2          & 39.5            & 53.8   & 41.4              & 58.9             \\
   & DKRL(CNN) + RotatE & 40.6          & 54.8          & 41.3            & 57.4   & 39.8              & 60.1             \\
   & DRKA(DPR) + TransE$_{k=5}$       & 38.2 & 51.4 & 37.3   & 53.9            & 38.8     & 58.1    \\ 
   & DRKA(DPR) + DistMult$_{k=5}$       & 35.7 & 49.0 & 36.6   & 52.1            & 37.0     & 56.5    \\ 
& DRKA + TransE$_{k=5}$       & \underline{41.2} & \textbf{55.7} & 39.7   & \underline{54.1}            & \underline{41.8}     & \underline{61.1}    \\ 
& DRKA + DistMult$_{k=5}$      & 40.3 & \underline{54.8} & 39.3   & 54.1            & 39.2     & 59.4    \\ 
& DRKA + CompIE$_{k=5}$      & 40.3 & 54.5 & \underline{40.2}   & 53.7            & 41.2     & 59.7    \\ 
& DRKA + RotatE$_{k=5}$      & \textbf{42.7} & \textbf{55.7} & \textbf{42.5}   & \textbf{58.9}            & \textbf{43.1}     & \textbf{63.8}    \\ 
\bottomrule
\end{tabular}
}
\end{minipage}
\caption{Link prediction results on the test split set on FB15K. The upper section includes results obtained in a KG standalone setup, where KGE models are used to learn from KG's alone; the lower section results are obtained when the KGE models are augmented using textual descriptions, as covered in Section \ref{sec:text-kg} (Text-enhanced Knowledge Graph Completion). DRKA(DPR) Best and second-best results are formatted respectively as bold and underlined text, across each column for the KG and KG+Text setup.}
\end{center}
\label{tab:main results}
 \vspace{-2em}
\end{table}

\subsection{Training}
We initialise four different KGE learning models (as Section \ref{sec:baseline} indicates), and use them to extract KG embeddings, and initialise SBERT for text embeddings. The number of negative samples per triple is set to 100 and $k$ is set to 5. We tune all hyper-parameters using the validation data, and obtain optimal values as follows: learning rate - 1e-3, batch size - 8, KG embedding size - 200. Further details on tuning bounds are provided in Table~\ref{tab:parameter}.

\subsection{Evaluation Results}
\label{sec:eval_results}
\paragraph{Setup:} We perform two sets of experiments for the different evaluation tasks. Initially, we use the KGE models as stand alone methods, and we later test KGE embeddings that are augmented with text description embeddings. We consider scenarios with triples whose entities co-occur within the text descriptions (With mentions), triples whose entities do not co-occur within the text descriptions (Without mentions) as well as all of the triples together (Overall).  

\paragraph{Link Prediction (LP):}
LP is a popular KG completion task which attempts to evaluate how well KGE, text-enhanced KGE and Pre-trained LMs predict either a missing subject entity from a given triple $(?,r,e_o)$, or a missing object entity from a triple $(e_s,r,?)$. Table 3 shows performance results for the various models across the two different setups. We observe RotatE outperforming the other KGE across all 6 experiments in the KG only setup. However, we notice both sets of models perform slightly better with triples whose elements are not mentioned in text (without mentions), compared to those with textual mentions. 

\begin{table}[!b]
\begin{center}
\begin{tabular}{@{}llcccc@{}}
\toprule
          &                       & \multicolumn{2}{c}{Overall} & \multicolumn{2}{c}{Without mentions} \\ 
          &                       & MR  $\downarrow$         & Hits@1 $\uparrow$          & MR  $\downarrow$              & Hits@1  $\uparrow$              \\ \midrule
KG Only   & TransE                & 4.2          & 86.7         & 4.1              & 87.3              \\
          & DistMult              & 4.7          & 85.4         & 4.8              & 85.0              \\
          & CompIEx            & \underline{4.0}         & \underline{86.9}         & \underline{4.0}              & \underline{88.1}              \\
          & RotatE              & \textbf{3.5}          & \textbf{89.0}         & \textbf{3.1}              & \textbf{89.5}              \\\midrule
KG + Text & DKRL(CNN) + TransE   & 2.9          & 88.1         & 2.7              & 89.4              \\
          & DKRL(CNN) + DistMult & 3.1          & 87.6         & 3.0              & 88.1              \\
           & DRKA(DPR) + TransE$_{k=5}$         & 3.2          & 84.2        & 3.1              & 85.8             \\
            & DRKA(DPR) + TransE$_{k=5}$         & 3.7          & 82.7         & 3.3             & 84.1            \\
          & DRKA + TransE$_{k=5}$         & \underline{1.9}          & 91.5       & \underline{1.7}              & \underline{92.2}             \\
          & DRKA + DistMult$_{k=5}$       & 2.2          & 90.3         & 1.9              & 90.8              \\
          & DRKA + CompIEx$_{k=5}$       & 1.8          & \underline{91.8}         & \underline{1.7}             & 91.6  \\
          & DRKA + RotatE$_{k=5}$       & \textbf{1.1}          & \textbf{93.4}         & \textbf{1.1}              & \textbf{93.7}  \\ \bottomrule
\end{tabular}
\label{tab:relation_prediction}
\caption{Relation prediction results (Mean Rank (MR) and Hits@1) for the KG Only and KG + Text setup on FB15K dataset. The lower($\downarrow$) the MR score, the better and the higher ($\uparrow$ ) the Hits@1, the better.}
\end{center}
 \vspace{-2em}
\end{table}

We notice that augmenting the KGE with text descriptions (KG + text setup) significantly improves the performance in all three scenarios, with and without mentions as well as overall. DRKA+RotatE produces the best results in majority of the experiments (5/6 to be precise) followed by DRKA+TransE. We observe the best performing model DRKA + RotatE$_{k=5}$ outperform the baseline architecture DKRL(CNN) + RotatE i.e. the average percentage increase in MRR and Hits@10 across the three setups is 5.5\% and 3\% respectively. We attribute this performance to the fact that DRKA injects richer, semantically relevant contextualised representations (of the text descriptions) obtained by a transformer-based (SBERT) retriever model. On the otherhand, DKRL(CNN)+TransE injects word2vec representations initialised using word2vec. While word2vec and other traditional word embedding models are context-insensitive (i.e. a word embedding is fixed irrespective of the context in which its mentioned), contextualised embedding models such as BERT dynamically produce word embeddings, in other words, different contexts trigger different embeddings for the same word. This ultimately enhances the learning of different meanings and senses \shortcite{loureiro2021analysis} in language modelling tasks, such as those covered in this work.  
We additionally observe DRKA(DPR) models perform poorly in comparison to the other models, and we attribute this to error propagation as a result of decoupling the retrieving from the alignment process i.e. errors that originate from the retrieving process done by DPR automatically affect the alignment process. Furthermore, the query formulation process is simply a concatenation of the triple elements, rather than an actual question with elaborative context about the triple elements.

\paragraph{Relation Prediction (RP):}
Similar to LP, RP aims to predict a missing element of a triple, however RP specifically looks to predict a missing relation from $(e_s,?,e_o)$. Table 4 shows RotatE and DRKA + RotatE dominating the performance in the KG stand alone setup and KG + Text setup respectively. These results prove that supplementing the structured KGE with text can leas to significant performance gains in gains in not just LP, but RP too. 

\paragraph{Triplet Classification (TP):} 
Similar to prior work, we define TP as a binary classification task which classifies an entity triple $e_t$ as a valid or invalid triple. We adopt the evaluation protocol used by \shortciteA{wang2014knowledge} when generating negative samples i.e. we construct a false triple by corrupting a valid KG triple. For $(e_s,r,e_o) \in KG$, where $\{e_s, e_r\} \in \mathcal{E}$ we 1) 
replace $e_s$ with a random entity $e_s^{'}$ 2) replace $e_o$ with a random entity $e_o^{'}$ and 3) Replace both subject and object entities with random entities, where $\{e_s^{'},e_o^{'}\} \in \mathcal{E}$. We further repeat steps 1 to 3 , yet this time sampling the corrupt entities from the text corpus.

Table 5 shows that the models struggle less in predicting validity of valid triples, as seen in column two $e_s,r,e_o$ i.e. the models perform best in contrast to all the other triple types investigated.

\begin{table}[t]
\begin{center}
\begin{tabular}{@{}lcccc@{}}
\toprule
                 & $e_s,r,e_o$ & $e_s^{'},r,e_o$ & $e_s,r,e_o^{'}$ & $e_s^{'},r^{'},e_o^{'}$ \\ \midrule
TransE           & 91.2        & 53.8        & 55.4        & 81.1        \\
DistMult         & 89.1        & 49.8        & 55.5        & 80.6        \\
DRKA + TransE    & \textbf{95.7}        & 60.4        & 63.2        & \textbf{84.5}        \\
DRKA + DistMult  & 94.2        & \textbf{61.5}       & \textbf{63.3}        & 83.2        \\ \midrule
TransE*          & 91.2        & 58.2        & 60.2        & 83.6        \\
DistMult*        & 89.1        & 54.3        & 59.1        & 82.7        \\
DRKA + TransE*   & \textbf{95.7}        & 66.8        & 65.6        & \textbf{88.6}       \\
DRKA + DistMult* & 94.2        & \textbf{70.2}        & \textbf{72.7}        & 88.5        \\ \bottomrule
\end{tabular}
\caption{Triplet classification accuracy (\%) over various types of triples. * indicates that the corrupted entities are drawn from the text corpus, rather than from the KB from which the KG is constructed. Only TransE and DistMult are tested for these experiments.}
\end{center}
\label{tab:triplet_classification}
 \vspace{-1em}
\end{table}

It is noticeable that corrupting the subject or head entity $e_s$ (in column 3) causes a bigger drop in performance compared to when the object or tail entity $e_o$ is corrupted (in column 4). The models perform relatively well when tasked with classifying triples with invalid entities and relations (in column 5), despite performing worse with valid triple types.
We also observe that sampling negative or corrupt entities from the corpus does not lead to the same performance deterioration as it does when they are sampled from the KG. This is attributed to fact that the retriever model selects a core set of related text descriptions that are relevant to a triple and hence enhancing the models ability to detect presence or absence of a triple within the text.

\begin{figure}[t]
    \centering
    \includegraphics[width=0.5\columnwidth]{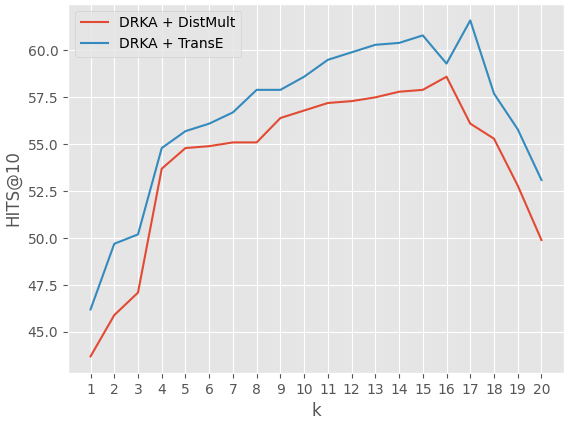}
    \caption{Adjusting $k$ to determine how many text descriptions would be relevant in augmenting an entity representation for the Link Prediction task.}
    \label{fig:dkra_k}
\end{figure}

\subsection{Ablation and Analysis}
To further understand the performance of our proposed multi-task framework, we conduct a set of two other investigations. The first involves adjusting parameter $k$ which specifies the number of text descriptions to use in augmentation, and the second involves eliminating the retriever component and simply use a randomly selected set of relevant text descriptions. These two investigations are detailed in the following subsections.

\subsubsection{Adjusting $k$}
To determine the optimal number of relevant descriptions as well as compare using a single text description in augmenting KG embeddings, we tune the $k$ value (for the LP task) within the range $\{1,20\}$ where $1$ and $20$ are the lower and upper bounds, respectively. The lower bound ultimately represents augmentation premised on a single description whereas, all $k \ge 1$ settings represent augmentation premised on multiple text descriptions.
Figure \ref{fig:dkra_k} illustrates the significant improvement in performance when the number of relevant text descriptions ($k$) to use in the augmentation process is increased from 1 on wards. The figure shows a sharp rise in Hits@10 performance through from the first 5 $k$ values, which is followed by a gradual increase all the way to $k=16$. The performance is seen to decline in the last 5 $k$ values. We attribute this to a possible overlap in descriptions for triples, which then confuses the model at inference. This decline is an indicator that, despite the gains made by using more descriptions, a high number of descriptions $k > 16$ may instead hurt the performance of the text-enhanced KG embedding model in tasks such as LP.

\subsubsection{DRKA - Retriever}
We obviate the role of the retriever and instead arbitrary select a set of entity descriptions existing in the Babelnet corpus \shortcite{navigli2012babelnet}, Google News dataset, and Wikipedia articles. The selected or targeted descriptions are still relevant to the entities, however, they are obtained in a distantly supervised manner, where we simply search and select a description provided for a given entity. For a given triple $(e_s,r,e_o)$, two pairs of relevant descriptions are randomly selected for the subject $e_s$ and object $e_o$ respectively. We chose the same number of descriptions (2) per entity to avoid skewing the semantics captured by embedding towards one entity and instead allow a proportionate distribution across both featuring entities. 

These descriptions are aligned to the triple embedding $\vec{e_t}$ (obtained by concatenating triple elements) following the same procedure (under subsection \ref{sec:text-entity_fusion}) of computing an attention scores of these 4 documents to the triple and then generating an overall representation as an inner product between the attention scores and the description embeddings projected into the KG vector space. This setup however eliminates the retriever loss, thereby only minimising the alignment loss $L_{align}$. 

\begin{table}[h!]
\begin{center}
\begin{tabular}{@{}lcccc@{}}
\toprule
& \multicolumn{2}{c}{Link Prediction} & \multicolumn{2}{c}{Relation Prediction} \\ 
& MRR $\uparrow$           & Hits@10 $\uparrow$         & MR $\downarrow$            & Hits@1 $\uparrow$                \\ \midrule
DRKA + TransE$_{k=4}$   & \textbf{40.7}        & \textbf{54.8}       & \underline{1.9}           & \textbf{90.7}           \\
DRKA + DistMult$_{k=4}$   & 40.1        & \underline{53.7}        & 2.3          & \underline{90.3}           \\ \midrule
DRKA - retriever + TransE$_{k=4}$   & 39.8        & 51.5        & \underline{1.9}          & 89.7  \\
DRKA - retriever + DistMult$_{k=4}$ & 37.9        & 51.2        & 2.0           & 87.4           \\ \midrule
DRKA - retriever + TransE$_{k=6}$   & \underline{40.9}        & 52.1        & \textbf{1.3}           & 89.3           \\
DRKA - retriever + DistMult$_{k=6}$ & 39.1        & 51.9        & 1.8           & 88.4           \\ \bottomrule
\end{tabular}
\label{tab:drka-retriever}
\caption{Link and Relation prediction results of DRKA with and withorugh the retriever component (- retriever) for two different values of $k$, ie. $k=4$ and $k=6$. The best and second best scores are in bold and underlined respectively.}
\end{center}
 \vspace{-2em}
\end{table}

Table 6 shows results of the multi-task DRKA frame work trained without the retriever component and instead a set of descriptions is selected from the corpus randomly but still relevant to the entities in a given triple as described in the preceding paragraphs. For consistency, we re-run the full DRKA framework with $k$ set to 4 because its the same number considered in the DRKA - retriever setup. As shown in the table, there is a drop in performance when the retriever is deducted, more so, a significant drop in Hits@10. This drop is further evidence of the significance of a model trained to explicitly select a set of descriptions that are semantically relevant to the triple of entities. 

To probe this impact further, we additionally test using $k=6$, selecting 3 descriptions per entity. We hypothesize that an increase in the number of descriptions might subtly eliminate the need of a retriever model. On the contrary, we realise that the performance still drops however, to a degree lesser than it does in the $k=4$ setting. The DRKA - retriever + TransE$_{k=6}$ achieves the best Mean rank score in the RP task. These changes indicate that select an appropriate number to use in augmenting is so critical and can have a good or detrimental impact on the performance of the model.

\section{Conclusion}
This paper has explored dense representation learning as a conduit for achieving KG augmentation. It proposes a retriever based augmentation model, called DRKA, that jointly learns KG embeddings and contextualised embeddings of text produced by a dense representation model, SBERT. The initial set of evaluation experiments performed showed that augmenting KG embeddings with dense representations of text descriptions using DRKA improves performance in KG completion tasks such as Link Prediction, Relation Prediction and Triplet Classification. We are aware of the significant impact that text-enhanced KGE have had in KG completion, however, this paper has shown that increasing the number of text descriptions to use in augmenting KG embeddings can lead to further gains in performance of these models. Having said that, we observed that constantly increasing the number of descriptions to incorporate into the KG embedding may at a point, begins hurting the performance of the model. For further analysis, we investigate the impact of the retriever component within DRKA i.e. train and evaluate DRKA minus the retriever. We discover that deduction of the retriever hurts the performance of the model. Overall, this paper has empirically demonstrated that enhancing KGE models with semantically rich dense representations of text can benefit KG completion. Its proposed framework can be adapted to domain specific KG tasks and can eliminate the need to supply a set entity descriptions manually collated from multiple sources i.e the model is able to automate retrieval of relevant text descriptions to use for augmenting. 

\vskip 0.2in
\bibliography{article}
\bibliographystyle{theapa}

\end{document}